\documentclass{article}
\usepackage{spconf,amsmath,graphicx,hyperref}

\usepackage{cite}
\usepackage{amsmath,amssymb,amsfonts}
\usepackage{graphicx}
\usepackage{textcomp}
\usepackage{xcolor}
\def\BibTeX{{\rm B\kern-.05em{\sc i\kern-.025em b}\kern-.08em
    T\kern-.1667em\lower.7ex\hbox{E}\kern-.125emX}}

\usepackage{graphics} 
\usepackage{epsfig} 
\usepackage{mathptmx} 
\usepackage{booktabs}
\usepackage{multirow}
\usepackage{multicol}
\usepackage{times} 

\usepackage[dvipsnames]{xcolor}
\usepackage{algorithm}
\usepackage{algpseudocode}
\usepackage{algorithmicx}
\usepackage{float}
\usepackage{balance}

\algdef{SE}[DOWHILE]{Do}{doWhile}{\algorithmicdo}[1]{\algorithmicwhile\ #1}%

\algrenewcommand\algorithmicrequire{\textbf{Input:}}
\algrenewcommand\algorithmicensure{\textbf{Output:}}

\title{Privacy-Concealing Cooperative Perception for BEV Scene Segmentation}
\threeauthors
  {Song Wang\thanks{This work was partly supported by NSFC No. 62301497, the Science and Technology Research Program of Henan, No. 252102211024, the Key Research and Development Program of Henan, No. 231111212000, the Natural Sciences and Engineering Research Council of Canada (NSERC), and the Rogers Cybersecure Catalyst Fellowship Program.}}
	{\normalsize School of Elec. and Info. Engineering\\
    \normalsize Zhengzhou University\\
	\normalsize  Zhengzhou, China}
  {Lingling Li}
	{\normalsize School of Computer Science\\
    \normalsize  Zhengzhou University of Aeronautics\\
    \normalsize  Zhengzhou, China}
  {Marcus Santos, Guanghui Wang\sthanks{Corresponding author (Email: wangcs@torontomu.ca)}}
 {\normalsize Department of Computer Science\\
 \normalsize  Toronto Metropolitan University\\
	\normalsize  Toronto, Canada}
\begin{document}
%
\maketitle
\begin{abstract}
Cooperative perception systems for autonomous driving aim to overcome the limited perception range of a single vehicle by communicating with adjacent agents to share sensing information. While this improves perception performance, these systems also face a significant privacy-leakage issue, as sensitive visual content can potentially be reconstructed from the shared data. In this paper, we propose a novel Privacy-Concealing Cooperation (PCC) framework for Bird’s Eye View (BEV) semantic segmentation. Based on commonly shared BEV features, we design a hiding network to prevent an image reconstruction network from recovering the input images from the shared features. An adversarial learning mechanism is employed to train the network, where the hiding network works to conceal the visual clues in the BEV features while the reconstruction network attempts to uncover these clues. To maintain segmentation performance, the perception network is integrated with the hiding network and optimized end-to-end. The experimental results demonstrate that the proposed PCC framework effectively degrades the quality of the reconstructed images with minimal impact on segmentation performance, providing privacy protection for cooperating vehicles. The source code will be made publicly available upon publication.
\end{abstract}
\begin{keywords}
Privacy-concealing cooperation, adversarial learning, BEV, semantic segmentation
\end{keywords}

\section{Introduction}
\label{sec:intro}
Autonomous vehicles (AVs) are expected to generate a robust and safe driving plan by perceiving the surrounding environment \cite{chen2024end}. Recent deep learning-based techniques have made remarkable progress in perception tasks, such as semantic segmentation \cite{wang2025entroformer,zhao2024maskbev}, 3D object detection \cite{li2025pf3det,you2024mambabev}, classification \cite{li2022robust}, and occupancy prediction \cite{hou2024fastocc}. However, the perception capabilities of individual vehicles are limited by their short visual range. This limitation is particularly challenging in occlusion scenarios \cite{han2023collaborative}, where AVs struggle to make optimal driving decisions due to an incomplete understanding of their surroundings.

\begin{figure}[t]
  \centering
  \includegraphics[width=1\linewidth]{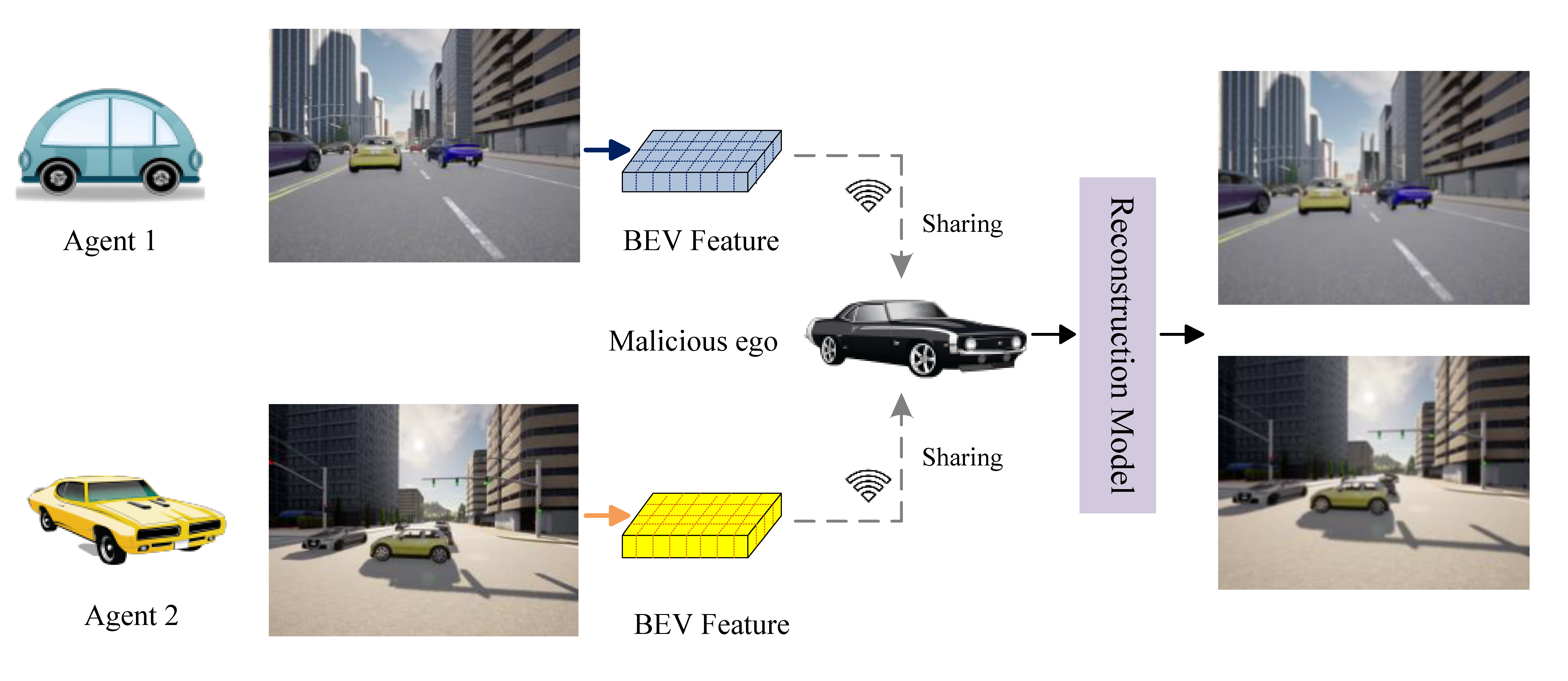}

   \caption{Reconstruction from BEV features by a malicious ego.}
   \label{fig:requirement}
\end{figure}

To overcome the limitation of short-range perception, multi-agent cooperation techniques offer a promising solution for long-range perception. These techniques aim to extend the perceptive range of individual vehicles by enabling communication and information sharing among nearby agents \cite{xu2022v2x}. However, limited bandwidth for communication between ego vehicles and neighboring agents presents a bottleneck that hinders the enhancement of perception performance in real-world multi-agent cooperation systems. While the raw data captured by agents contain the richest information for perceiving surrounding objects, transmitting such data requires enormous bandwidth resources \cite{han2023collaborative}. To address this issue and balance the performance-bandwidth trade-off, Bird’s Eye View (BEV) features have become the most popular choice for sharing and fusion in multi-agent cooperation systems \cite{hu2022where2comm}. Compared to Perspective View (PV), which suffers from occlusion and scale issues \cite{li2023delving}, BEV offers an intuitive representation of the world in which objects with semantic information can be accurately localized on a fixed-scale map, facilitating the integration of information from other agents \cite{ma2024vision}. Moreover, BEV features are significantly smaller in size than raw data, easily meeting the requirements for real-time communication. As a result, BEV features are considered the default for sharing in most current multi-agent cooperative perception techniques \cite{xu2022cobevt,lu2024an}.

\begin{figure*}[t]
  \centering
  \includegraphics[width=0.8\linewidth]{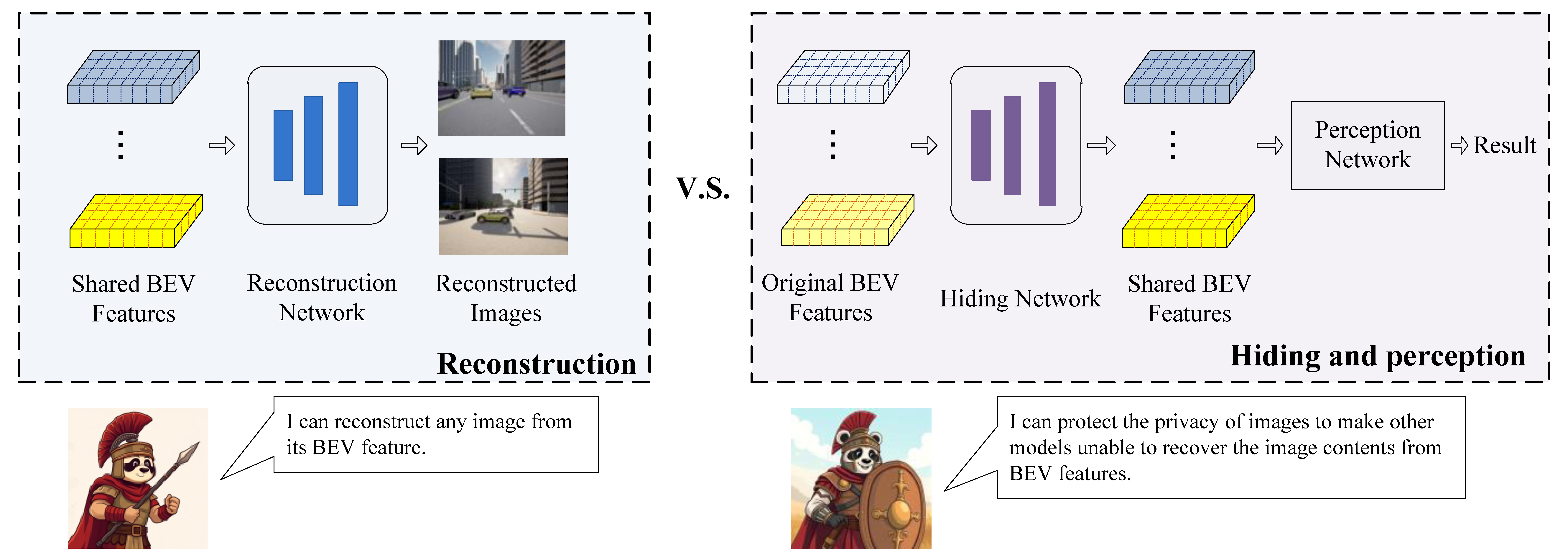}

   \caption{The proposed privacy-concealing cooperation framework. The reconstruction network aims to recover as much detail as possible from the received BEV feature maps and their corresponding images. The hiding network works to prevent the reconstruction network from recovering the image content while still preserving the perception performance.}
   \label{fig:ppc}
\end{figure*}

Despite advancements in perception performance, information sharing in multi-agent cooperation systems presents a significant risk of privacy leakage. Ideally, the shared BEV features should not reveal any image content while still contributing effectively to perception tasks. However, due to the powerful representational abilities of neural networks, recent research has demonstrated that image content can be accurately reconstructed from even local point features \cite{ng2022ninjadesc}. Since the shared BEV features in multi-agent cooperation systems typically contain rich surrounding information for perception tasks, they can easily be used to reconstruct the original perspective-view images, which may include sensitive objects. As illustrated in Fig. \ref{fig:requirement}, details such as the type and color of vehicles can be recovered from shared BEV features for tracking purposes. In this scenario, criminals could potentially track specific vehicles, including police cars, without being physically present. The shared features from nearby agents could become a powerful tool for locating targeted objects. Therefore, a robust cooperative perception system, when deployed in the real world, must carefully balance both high perception performance and stringent security considerations.

In this paper, we propose a Privacy-Concealing Cooperation (PCC) framework to prevent the leakage of visual content while maintaining high-perception performance. Specifically, we design a hiding network to encode the BEV features and conceal visual clues, while a reconstruction network is introduced to adversarially recover the image content from the encoded BEV features. To preserve perception performance, the downstream network, including BEV fusion and perception head, is retrained during the adversarial learning process. By allowing the hiding and reconstruction networks to compete with each other, PCC learns to prevent visual content reconstruction and mitigate performance degradation simultaneously. We validate the proposed PCC framework on the state-of-the-art (SOTA) cooperative BEV semantic segmentation model (CoBEVT \cite{xu2022cobevt}). Experimental results demonstrate that the BEV features encoded by the hiding network lead to only a negligible performance drop for the BEV semantic segmentation task, while significantly reducing the visual similarity between the reconstructed images and the original inputs.

\section{Method}
\label{sec_method}

This section outlines the details of the proposed PCC framework for BEV semantic segmentation. We first present the overall framework in Section \ref{sec_ppc}, followed by the reconstruction network in Section \ref{sec_rec}, and the privacy-concealing network in Section \ref{sec_hid}. The training strategy for the PCC framework is provided in Section \ref{sec_al}.

\subsection{Privacy-concealing cooperation framework}
\label{sec_ppc}
The proposed privacy-concealing cooperation framework is illustrated in Fig. \ref{fig:ppc}. The framework primarily includes two networks: the reconstruction network $R$ and the hiding network $H$. Given the received BEV feature maps, the reconstruction network aims to recover as much content detail as possible from the corresponding images. In contrast, the hiding network is responsible for protecting the image content from recovery by the reconstruction network while still maintaining perception performance. To achieve these goals, this paper employs an adversarial learning mechanism, enabling the networks to compete with each other and enhancing the privacy-concealing ability of the hiding network. Since the hiding network may alter the feature representation of perceived objects, the BEV perception network is retrained along with the hiding network to prevent a deterioration in perception performance. The motivation of designing adversarial networks to learn to protect the privacy of BEV features is the powerful representation ability of neural networks. Given that a reconstruction network is trained to recover the image content, it is natural that a hiding network can be trained to hide the visual information in BEV features. Therefore, we propose to {\it use the spear of the deep network to attack its shield}, degrading the quality of the reconstructed images.

\subsection{Reconstruction network}
\label{sec_rec}
The reconstruction network is designed to recover visual contents from the shared BEV features. For each agent, we denote the captured image as $I$. After the visual feature extraction, view transformation, and content protection, the shared BEV feature map $B_s$ is generated. Upon receiving $B_s$, the reconstruction network aims to reconstruct the image $\hat{I}$ from $B_s$. We employ the simple MAE decoder \cite{he2022masked} for the reconstruction network. The reconstruction process is formulated as:
\begin{equation}
  \hat{I} = R(B_s).
  \label{eq:recon}
\end{equation}

The reconstructed $\hat{I}$ is expected to be the same as the image $I$. To reach this goal, we build a reconstruction loss that is formulated as: 
\begin{equation}
  L_{Rec} = L_{dis}+\alpha L_{perc},
  \label{eq:l_rec}
\end{equation}
where $L_{dis}$ stands for the mean absolute error function to minimize the pixel-level distance between the reconstructed image $\hat{I}$ and the input image $I$. The perceptual loss $L_{perc}$ is used to improve the consistency of the content between $\hat{I}$ and $I$. In this paper, we employ the 16-layer VGG model \cite{simonyan2014very} pretrained on the ImageNet dataset \cite{russakovsky2015imagenet} to extract the semantic representation of the image contents. Since the reconstruction network primarily focuses on the fidelity of the reconstructed images to ground truth, we set $\alpha = 0.1$ in our experiments to emphasize $L_{dis}$ over $L_{perc}$.

\subsection{Hiding network}
\label{sec_hid}
To prevent the reconstruction network from recovering the visual contents from BEV features, we design a lightweight and plug-and-play hiding network for all the safe-aware cooperative perception systems. The hiding network contains six convolution layers with $1\times 1$ kernels, each followed by a ReLU layer, except for the last one. An InstanceNorm layer and a Dropout layer are inserted before the second and the fifth convolution layers, respectively. Given an original BEV feature map $B_o \in \mathbb{R}^{h\times w \times C}$, the hiding network conceals the visual information and converts it into a new feature map $B_s \in \mathbb{R}^{h\times w \times C}$ to share. The hiding process is formulated as follows.
\begin{equation}
  B_s = H(B_o).
  \label{eq:hiding}
\end{equation}

\begin{table*}[t]
\small
\caption{Quantitative results of image reconstruction attacks and retained perception utility on the OPV2V validation set. The threshold $T = 0.01$ in the metric PHV is set to be a margin to separate the reconstructed images and the ground truth. The arrows in the visual clue concealment column indicate that higher/lower values are better for privacy.}
\label{table_visualconceal}
\begin{center}
\scalebox{0.95}{
\begin{tabular}{|c||c|ccccc|c|c||c||c|}
\hline
\multirow{3}{*}{Methods} & \multicolumn{8}{|c||}{Visual clue concealment} & Map-view seg.&  Inference\\
\cline{2-11}
& \multirow{2}{*}{FID($\uparrow$)} &\multicolumn{5}{c|}{PHV$_{T=0.01}$($\uparrow$)} &\multirow{2}{*}{SSIM($\downarrow$)}  &\multirow{2}*{PSNR($\downarrow$)} & \multirow{2}{*}{IoU(\%)}& \multirow{2}{*}{Time(ms)}\\
\cline{3-7}
& & layer1 & layer2 & layer3 & layer4 & average& & & & \\
\hline
{\scriptsize COBEVT (Original BEV)} 
& 281.5 & 0.6861 & 0.6671 & 0.5308& 0.9635 & 0.7119 & 0.3524 & 13.4418 & 57.2913 & 158.75\\
\hline
{\scriptsize PCC (Privacy-concealed BEV)} 
& 378.1 & 0.7058& 0.7033 & 0.5685 &0.9673 & 0.7362 & 0.3471 & 12.8889& 57.2768& 174.23\\
\hline
\end{tabular}}
\end{center}
\end{table*}

The transformation of the BEV feature map significantly alters the explicit distribution of visual semantic information, causing the cooperative perception model to become invalid. To achieve performance comparable to the original model, this paper retrains the cooperative perception network $P$ during the adversarial learning process. The same cooperative architecture is used in retraining to avoid introducing additional computational overhead.

The hiding and cooperation models are optimized for specific perception tasks, such as segmentation and detection. In this paper, we use BEV semantic segmentation to validate the effectiveness of the proposed privacy-concealing cooperation framework. Therefore, the loss for the hiding and cooperation models is defined as:
\begin{align}
\label{eq:coop}
\begin{aligned}
    L_{Coop}  = L_{Seg}
    = CrossEntropy(P(B_s), GT),
\end{aligned}
\end{align}
where the cross entropy loss is used to measure the similarity between the predicted segmentation map and the ground truth (GT). 
The designed hiding network is highly lightweight, with only 214.4K parameters. When integrated into cooperative perception systems, it requires minimal computational resources to protect visual privacy during inference.

\subsection{Adversarial learning}
\label{sec_al}
The proposed privacy-concealing cooperation framework aims to hide the visual clues of the shared BEV feature maps by making the reconstruction network and the hiding network compete with each other. More specifically, $H$ is designed to maximize the reconstruction error, while the optimization objective of $R$ is to recover as many image content details as possible by minimizing the reconstruction error. This problem can be defined as the following minimax game.
\begin{equation}
  \min_R \max_H J(R,H) = L_{Rec}(B_o, I; R, H).
  \label{eq:minimax}
\end{equation}

To prevent deterioration in perception performance, the proposed privacy-concealing cooperation framework is also designed to minimize the cooperative perception loss as: 
\begin{equation}
  \min_{H,P}L_{Coop}(B_o, GT;H, P).
  \label{eq:minh}
\end{equation}

To solve the above problems, this paper separates them into two optimization objectives. First, for the reconstruction model, its objective is defined as: 
\begin{equation}
  \min L_{R}= L_{Rec}(B_o, I;R, H).
  \label{eq:obj_rec}
\end{equation}
Second, since the optimization of the hiding model $H$ is related to both Eq. (\ref{eq:minimax}) and Eq. (\ref{eq:minh}), we define its optimization loss as: 
\begin{equation}
  \min L_{H}= L_{Coop}(B_o, GT;H, P) - \lambda L_{Rec}(B_o, I;R, H),
  \label{eq:obj_hiding}
\end{equation}
where the parameter $\lambda$ adjusts the emphasis degree of $H$ on the visual content concealment over the perception performance. By optimizing Eq. (\ref{eq:obj_hiding}), we simultaneously reach the goals of concealing the visual clues and preserving perception performance.



\section{Experiment}
\label{sec_exp}

We adopt the SOTA cooperative BEV semantic segmentation model COBEVT \cite{xu2022cobevt} to validate the effectiveness of the proposed PCC framework on privacy concealing. We retrain the BEV perception network $P$ to maintain the segmentation performance after hiding the visual clue by $H$. Following the original setting of COBEVT, the OPV2V training set \cite{xu2022opv2v} is used to train the models $R$, $H$, and $P$, while their performance is evaluated on the validation set.

\subsection{Quantitative results}
We first conduct a quantitative analysis to assess the privacy-concealing ability of the hiding network and the retention capacity of the perception network for map-view segmentation performance. Taking the front image ``camera\_0'' of each vehicle as an example, multiple metrics, i.e. FID \cite{heusel2017gans}, PHV \cite{liu2021unpaired}, SSIM \cite{wang2004image}, and PSNR, are employed in this paper to measure the similarity between the reconstructed images and the input images.

Table \ref{table_visualconceal} presents the quantitative results of image reconstruction attacks and the retained perception utility on the OPV2V validation set. First, we compare the quality of images reconstructed from original BEV features and privacy-contained ones across multiple metrics. In the `Visual clue concealment' column of Table \ref{table_visualconceal}, the directions of the arrows next to each metric indicate the favorable trend for privacy. The table shows a significant drop in the quality of the reconstructed images after privacy concealment, demonstrating the effectiveness of the hiding model in protecting privacy. From the {`Map-view segmentation'} column of Table \ref{table_visualconceal}, we observe very close IoU value before and after concealment, indicating minimal impact on segmentation performance. This result confirms that the proposed PCC framework successfully conceals visual privacy in the shared BEV features while preserving perception accuracy.

\subsection{Qualitative results}

\begin{figure}[t]
      \centering
      \includegraphics[width=0.444\textwidth]{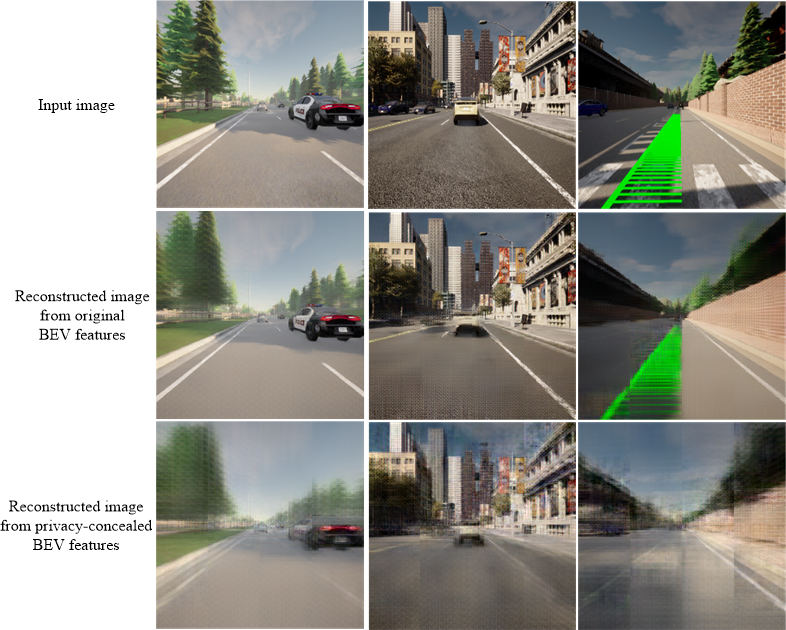}
      \caption{Qualitative results on the OPV2V validation set. }
      \label{fig_vis}
\end{figure}

To further validate the effectiveness of the proposed PCC framework in privacy concealment, we conduct a qualitative analysis in this section. Some qualitative results are shown in Fig. \ref{fig_vis}, where the first row presents input images from OPV2V, while the second and third rows show reconstructed images from the original BEV features and the privacy-contained features, respectively.  When comparing the images in the first two rows of Fig. \ref{fig_vis}, we observe clear visual details in the reconstructed images of the original BEV features, revealing the potential privacy leakage from reconstruction attacks. After adversarial learning, the visual details in the reconstructed images of the third row are significantly degraded, confirming the privacy-concealing capability of the proposed PCC framework.

\section{Conclusion}
In this paper, we have proposed a privacy-concealing cooperation framework for BEV semantic segmentation. Our approach employs an adversarial scheme in which the hiding network and the reconstruction network compete with each other. To preserve segmentation performance, we integrate the perception network with the hiding network and train them in an end-to-end manner. The evaluation results across multiple metrics demonstrate that our framework achieves strong privacy concealment for BEV features while minimally affecting the accuracy of the segmentation.


\end{document}